\documentclass[sigchi]{acmart}

\usepackage{booktabs} 
\usepackage{diagbox}
\usepackage{gensymb}
\usepackage{balance}

\AtBeginDocument{%
  \providecommand\BibTeX{{%
    \normalfont B\kern-0.5em{\scshape i\kern-0.25em b}\kern-0.8em\TeX}}}

\setcopyright{acmcopyright}
\copyrightyear{2019}
\acmYear{2019}
\acmDOI{10.1145/3340555.3355713}

\copyrightyear{2019}
\acmYear{2019}
\acmConference[ICMI '19]{2019 International Conference on Multimodal Interaction}{October 14--18, 2019}{Suzhou, China}
\acmBooktitle{2019 International Conference on Multimodal Interaction (ICMI '19), October 14--18, 2019, Suzhou, China}
\acmPrice{15.00}
\acmDOI{10.1145/3340555.3355713}
\acmISBN{978-1-4503-6860-5/19/10}

\begin{document}

\title{Exploring Emotion Features and Fusion Strategies for Audio-Video Emotion Recognition}

\author{Hengshun Zhou$^{*,1}$, Debin Meng$^{*,2}$, Yuanyuan Zhang$^1$, Xiaojiang Peng$^{\dagger,2}$, Jun Du$^1$, Kai Wang$^2$, Yu Qiao$^2$}
\authornote{Hengshun Zhou and Debin Meng contributed equally to this research. \\
$\dagger$ Xiaojiang Peng is the corresponding author. Email: xj.peng@siat.ac.cn}
\affiliation{%
	\institution{ 
		$^1$National Engineering Laboratory for Speech and Language Information Processing (NEL-SLIP), University of Science and Technology of China, P.R. China
	\\ $^2$ShenZhen Key Lab of Computer Vision and Pattern Recognition, SIAT-SenseTime Joint Lab, Shenzhen Institutes of Advanced Technology, Chinese Academy of Sciences, China
	}
}
\renewcommand{\shortauthors}{Hengshun Zhou and Debin Meng, et al.}

\begin{abstract}
  The audio-video based emotion recognition aims to classify a given video into basic emotions. 
  In this paper, we describe our approaches in EmotiW 2019, which mainly explores emotion features and feature fusion strategies for audio and visual modality. For emotion features, we explore 
  audio feature with both speech-spectrogram and Log Mel-spectrogram and evaluate several facial 
  features with different 
  CNN models and different 
 emotion pretrained strategies. For fusion strategies, we explore 
  intra-modal and cross-modal fusion methods, such as designing attention mechanisms to highlights important emotion feature, exploring feature concatenation and factorized bilinear pooling (FBP) for cross-modal feature fusion. 
 With careful evaluation, we obtain 65.5\% on the AFEW validation set and 62.48\% on the test set and rank second in the challenge.
\end{abstract}



\maketitle

\begin{figure*}[htp]
\includegraphics[width=1\textwidth]{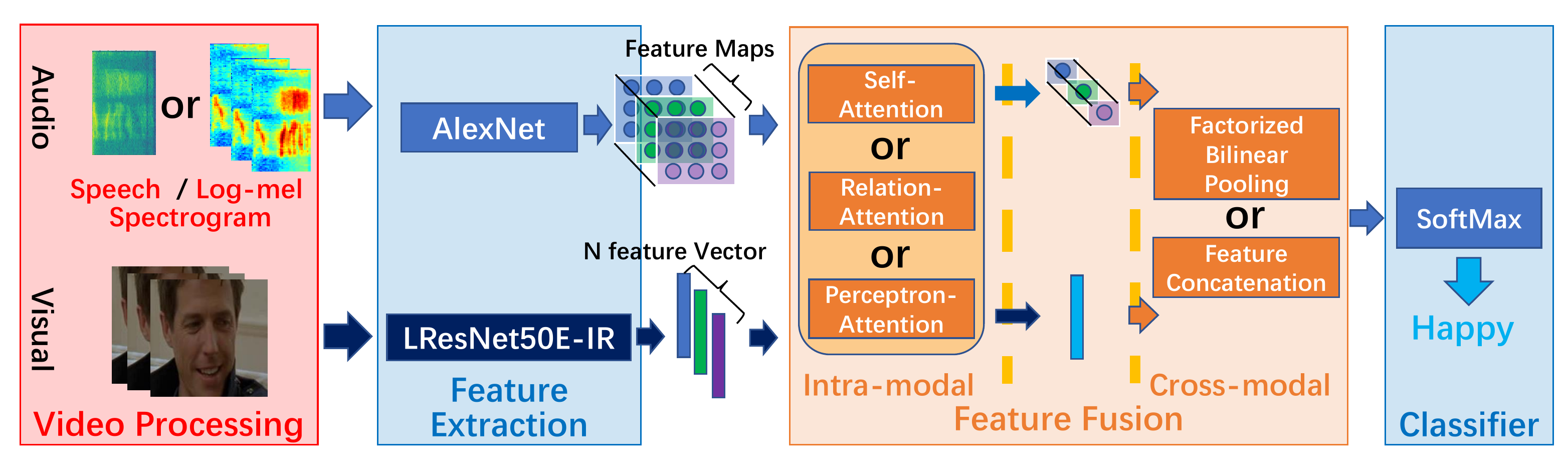}
\caption{The pipeline of audio-video emotion recognition.}
\label{fig:simple_pipeline}
\end{figure*}

\section{Introduction}
Emotion recognition(ER) has attracted increasing attention in academia and industry due to its wide range of applications such as human-computer interaction~\cite{dix2009human}, clinical diagnosis~\cite{mitchell2009clinical}, and cognitive science~\cite{johnson1980mental}. Although great progress in the face and video analysis has been made \cite{deng2018arcface,wen2016discriminative,tan2017group,wang2018deep,wang2019region,yang2018deep}, audio-video emotion recognition in the wild remains a challenging problem due to the expression 
suffers from the large pose, illumination variance, occlusion, motion blur, etc.

Audio-Video emotion recognition can be summarized as a simple pipeline shown in Fig~\ref{fig:simple_pipeline}, which includes four parts, namely Video preprocessing, Feature Extraction, Feature Fusion, and Classifier. Specifically, video preprocessing refers to extract the spectrogram of the audio, the faces or landmarks of video. Feature extraction and feature fusion 
respectively extracts emotion features from the audio or visual signal and fuses emotion features into compact feature vectors, which are subsequently fed into a classifier for prediction.

Reviewing the methods of Audio-Video emotion recognition, we find that some methods emphasize feature extraction and other methods emphasize feature fusion. Yao et al~\cite{yao2016holonet} construct Holonet as discriminative feature extraction, which combines residual structure~\cite{he2016deep} and CReLU \cite{Shang2016Understanding} to increase network depth and maintain efficiency. 
The EmotiW2017 winner team \cite{Hu2017Learning} gets robust feature extraction with Supervised Scoring Ensemble (SSE) which adds supervision to intermediate layers and shallow layers. Since SSE only uses high-level representations, Fan et al\cite{fan2018video} further improve SSE by utilizing middle feature maps to provide more discriminative features.  
These methods mainly use average pooling to obtain video-level representation from frame-level.

Many feature fusion strategies have been used in previous EmotiW challenges. \cite{fan2016video,Vielzeuf2017Temporal,lu2018multiple} extract CNN-based frame features and use LSTM\cite{gers1999learning} or BLSTM\cite{graves2005framewise} to fuse them. \cite{Bargal2016Emotion,knyazev2018leveraging,liu2018multi} use Statistical encoding module to aggregate frame features which compute the mean, variance, minimum, and maximum of the frame feature vectors. However, these methods ignore the importance of frames. Besides, all previous methods mainly apply score averaging or feature concatenation for audio-video fusion, which ignores the correlation between the features from different modalities. 

In this paper, we exploit three types of intra-modal fusion methods, namely self-attention, relation-attention, and transformer\cite{Vaswani2017Attention}. They are used to learn weights for frame features to highlight important frames. 
For cross-modal fusion, we explore feature concatenation and factorized bilinear pooling (FBP) \cite{zhang2019deep}. Besides, we evaluate different emotion features, including convolutional neural networks (CNN) for audio information with both speech-spectrogram and Log Mel-spectrogram and several facial 
features with different 
CNN models and different 
emotion pretrained strategies. Finally, we obtain 62.48\% and rank second in the challenge.

Our contributions and finds can be summarized as follows.
\begin{itemize}
\item We experimentally show that better face recognition CNN models 
and choosing suitable emotion datasets to further pretrain the face CNN models is important. 
\item We design three kinds of attention mechanisms for visual and audio feature fusion. 

\item We apply a Factorized Bilinear Pooling (FBP) for cross-modal feature fusion.
\end{itemize}

\section{The proposed method}
We develop our ER system based on the pipeline of Video preprocessing-Feature Extraction-Feature Fusion-Classifier.

\subsection{Video preprocessing}
\subsubsection{Face detection and alignment.} We apply face detection and alignment by Dlib toolbox\footnote{http://dlib.net/}. We extend the face bounding box with a ratio of 30\% and then resize the cropped faces to scale of $224 \times 224$. We do not apply face detection and alignment for AffectNet dataset, due to the face bounding box had been provided. For AFEW dataset, If no face is detected in the picture, the entire frame is passed to the network. \subsubsection{Audio processing and Spectrogram calculation.} For each audio, the speech spectrogram and log Mel-spectrogram extraction process is consistent with \cite{zhang2019deep} and \cite{chen20183} respectively. For speech spectrogram, we use the Hamming window with 40 msec window size and 10 msec shift. Finally, the 200-dimensional low-frequency part of the spectrogram is used as the input to the audio modality. As for log Mel-spectrogram, we calculate its deltas and delta-deltas.

\subsection{Feature Extraction}

\subsubsection{Visual Features} We apply three CNN backbones to extract facial emotion features, namely VGGFace, ResNet18, and IR50~\cite{deng2018arcface}. The dimensions are 4096, 512, and 512, respectively.

\subsubsection{Audio Feature} We extract the feature maps of the audio from the last Pooling layer of AlexNet. The size of a 3-dimensional feature map is $H\times W\times C$, where the $H$($W$) is the height(width) of the feature map, and $C$ is the number of the channel of the feature map. The feature maps are then split into n vectors($n = H \times W$). Each vector is C-dimensional.

\subsection{Intra-modal Feature Fusion}
We apply the attention-based strategies for intra-modal feature fusion. It converts a variable number of emotion features(from audio or visual modality) into a fixed-dimension feature. We explore three attention methods, namely Self-attention, Relation-attention, and Transformer-attention.
Formally, we denote a number of emotion features as \{$f_1,\cdots, {f_n}$\}.

\subsubsection{Self-attention} We apply 1-dimensional Fully-Connected(FC) layer~$\mathbf{W_{d\times1}^0}$ and a sigmoid function $\sigma$ for each emotion feature, the weight of the $i$-th feature $f_i^T $ is defined by: 

\begin{equation}
\label{eq:alpha}
\alpha_{i}  = \sigma(f_i^T \cdot \mathbf{W_{d\times1}^0})
\end{equation}

With these self-attention weights, we aggregate all the emotion features into a global representation $f_s$ as follows:
\begin{equation}
f_s= \frac {\sum_{i=1}^{n}\alpha_{i}f_{i}}{\sum_{j=1}^{n}\alpha_{j}}.
\end{equation}

\subsubsection{Relation-attention} This attention module was designed to learn weights from the relationship between features. After the self-attention, features are aggregated into a single vector$f_s$. Since $f_s$ inherently contains global representation of these features, we use the sample concatenation of individual features and global represenation $[f_i : f_s]$ to model the global-local relation. Similar to the Self-attention module, with individual emotion features, we apply 1-dimensional FC layer~$\mathbf{W_{d\times1}^1}$ and a sigmoid function $\sigma$. The relation-attention weight of the $i$-th feautre $[f_{i}:f_s]^T$ is formulated as follows:

\begin{equation}
\beta_{i} = \sigma([f_{i}:f_s]^T \cdot \mathbf{W_{d\times1}^1}),
\end{equation}

With Self-attention and Relation-attention weights, all the emotion features was convert into a new feature as follows:
\begin{equation}
\label{eq:alphabeta}
f_r = \frac {\sum_{i=0}^{n}\alpha_{i}\beta_{i}[f_{i}:f_s]}{\sum_{j=0}^{n}\alpha_{j}\beta_{j}}.
\end{equation}

\subsubsection{Transformer-attention} Inspired by the works in\cite{zhang2019deep} and \cite{yang2016hierarchical}, we formulate the attention weight as follows:
\begin{equation}
\label{eq:transform}
f'_i = \mathbf{W_{m \times d}^2} \cdot f_i + b
\end{equation}
\begin{equation}
\label{eq:perceptron-attention}
\gamma_i = exp(\mathbf{u^t}_{m \times 1} \cdot tanh(f'_i))
\end{equation}

To reduce the dimension of the feature $f_{i}$, we use a $w \times d$-dimensional FC layer $\mathbf{W_{m \times d}^2}$ in Eq.(\ref{eq:transform}). Then the weight of the $i$-th feautre $f_{i}$ is processed by a $1$-dimensional FC layer $\mathbf{u^t}$, $exp()$ and $tanh()$ function in Eq.(\ref{eq:perceptron-attention}).

With these transformer-attention weights, we aggregate all the emotion features into a single feature $f_t$ as follows:
\begin{equation}
f_t= \frac {\sum_{i=1}^{n}\gamma_{i}f_{i}}{\sum_{j=1}^{n}\gamma_{j}}.
\end{equation}

\subsection{Cross-modal Feature Fusion} 
\begin{figure}[!ht]
\centering
\includegraphics[width=0.6\linewidth]{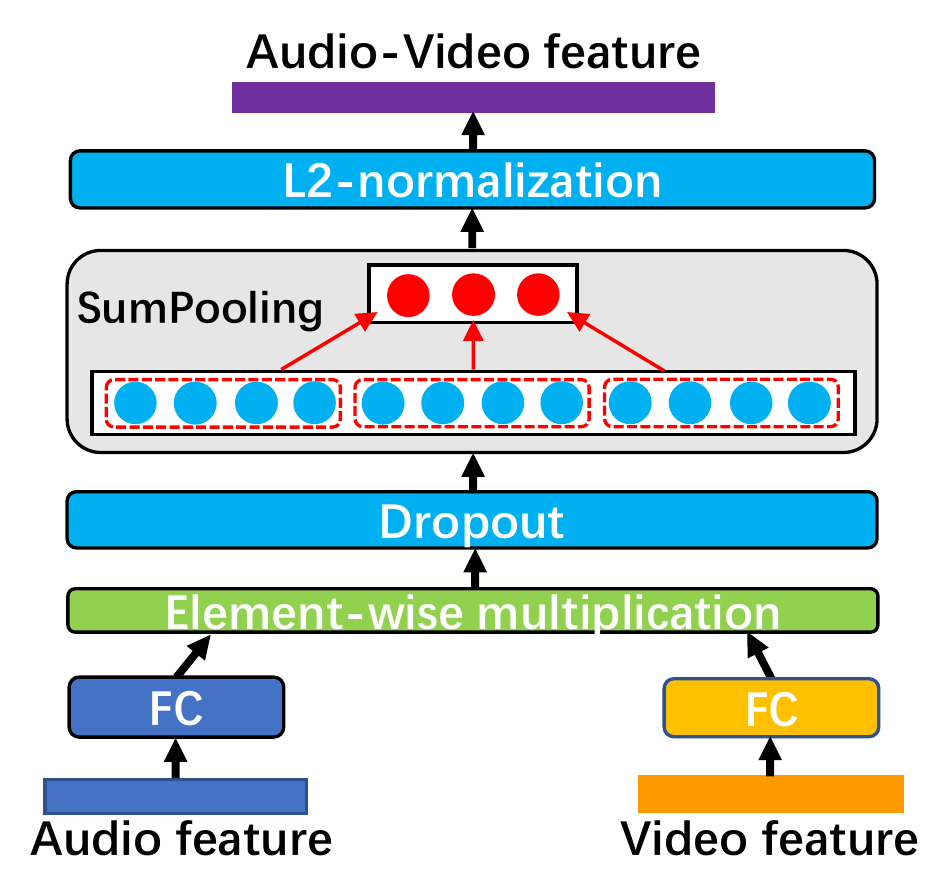}
\caption{Our factorized bilinear pooling(FBP) module.}
\label{fig:audio_video_FBP}
\end{figure}
We apply \textbf{Factorized Bilinear Pooling(FBP)} for cross-modal feature fusion. Given two features in different modalities,i.e. the audio feature vector $\boldsymbol{a}\in\mathbb{R}^m$ for a spectrogram and visual feature $\boldsymbol{v}\in\mathbb{R}^n$ for frame sequence, the simplest cross-modal bilinear model is defined as follows:
\begin{equation}
\label{bilinear_pooling}
{z}_{i}=\boldsymbol{a}^{T}\boldsymbol{W}_{i}\boldsymbol{v}
\end{equation} 

where $\boldsymbol{W} \in \mathbb{R}^{m\times n}$ is a projection matrix, $\boldsymbol{z_i} \in \mathbb{R}$ is the output of the bilinear model. we use the Eq.(\ref{sumpooling}) to obtain the output feature $\boldsymbol{z}=[z_{1},\cdots,z_{o}]$. The formula derivation from formula Eq.(\ref{bilinear_pooling}) to Eq(~\ref{sumpooling}) was discribed in the paper\cite{zhang2019deep}.
 \begin{equation}
 \label{sumpooling}
 \boldsymbol{z}=[z_{1},\cdots,z_{o}]={\rm SumPooling} (\tilde{\boldsymbol{U}}^{T}\boldsymbol{a}\circ\tilde{\boldsymbol{V}}^{T}\boldsymbol{v},k)
 \end{equation} 

The implementation of Eq(~\ref{sumpooling}) is illustrated in Fig\ref{fig:audio_video_FBP}, where $\tilde{\boldsymbol{U}}^{T}\boldsymbol{a}$ and $\tilde{\boldsymbol{V}}^{T}\boldsymbol{v}$ are implemented by feeding feature $\boldsymbol{a}$ and $\boldsymbol{v}$ to FC layers, respectively, and the function ${\rm Sum}{\rm Pooling}(\boldsymbol{x},k)$ applies sum pooling with non-overlapped windows to $\boldsymbol{x}$. Besides, Dropout is adopted to prevent over-fitting. The $l2$-normalization ($\boldsymbol{z}\leftarrow \boldsymbol{z}/\|\boldsymbol{z}\|$) is used to normalize the energy of $\boldsymbol{z}$ to avoid the dramatical variation of the output magnitude, due to the introduced element-wise multiplication. 

\section{EXPERIMENTS}
\subsection{Dataset}
In this work we use four emotion datasets to train our models, i.e. AffectNet\cite{Mollahosseini1949AffectNet}, RAF-DB\cite{li2019reliable}, FER+\cite{BarsoumICMI2016}, AFEW\cite{dhall2019emotiw,Dhall2012Collecting}. 

The human-annotated part of AffectNet dataset contains 287,651 training images and 4,000 test images, which are annotated with both emotion labels and arousal valence values. Only emotion labels are used in this task.

The RAF-DB dataset consists of 15,339 images labeled with 7-class basic emotion and 3,954 labeled with 12-class compound emotion. Only images labeled with basic emotion are used in this study. 

The FER+ dataset contains 28,709 training, 3,589 validation and 3,589 test images. We combine its training data with validation data for the training split and evaluate the model performance on the test data. 

The AFEW contains 773 train, 383 val and 653 test samples, which are collected from movies and TV serials with spontaneous expressions, various poses, and illuminations.

\subsection{Exploration of Emotion Features}
We explore emotion features in two perspectives, namely CNN backbones and pretraining emotion datasets.

For the choice of the CNN model, we compare IR50\cite{deng2018arcface}, ResNet18\cite{he2016deep}, and VGGFace\cite{parkhi2015deep} in the Table~\ref{tab:vgg_IR50_Res}, where the former two models are pretrained on MS-Celeb-1M dataset and the last one on VGGFace dataset. 
We find that the large CNN, IR50, is superior to the other two models.

We use the well-trained IR50 model to extract features and only train softmax classifier using these features. The IR50 models pre-trained on FER+, RAF-DB, and AffectNet achieve 50.13\%, 51.436\%, and 53.78\%, respectively. 
Therefore, we choose the IR50 model pretrain on AffectNet as our visual features in the following fusion experiments.

\begin{table}[htp]
\footnotesize
\centering
\caption{Exploration of CNN models and pretrained emotion datasets. 
}
\label{tab:vgg_IR50_Res}
\begin{tabular}{c|ccc}

\toprule
Model  & FER+  & RAF-DB  & AffectNet             \\ \midrule
VGGFace    &   88.84\%        & 86.93\%  & 51.425\%        \\ \midrule
ResNet18   &   88.65\%         & 86.696\%  & 52.075\%  \\

\midrule
IR50   & \textbf{89.257\% }   & \textbf{89.075\%}  & \textbf{53.925\% } \\
\bottomrule
\end{tabular}
\end{table}

\subsection{Exploration of Fusion Strategies}
We explore three intra-modal attention strategies with the FBP cross-modal fusion. We use speech spectrogram for audio CNN, which obtains 38\% on AFEW validation set individally. 
In the Table~\ref{tab:self_relation_perceptron}, we find the FBP improves performance for all the intra-modal fusion methods. Transformer attention for intra-modal fusion is the best for FBP.

\begin{table}[htp]
\footnotesize
\centering
\caption{Evaluation of intra-modal fusion methods. }
\label{tab:self_relation_perceptron}
\begin{tabular}{c|ccccc}
\hline
\diagbox{Audio}{Visual} & Self & Relation & Transformer\\ 
\hline
Self  & 54.6\% &     56.9\%     & 60.3\%      \\ \midrule
Relation    &   54.0\%     & 57.2\%     & 60\%        \\ \midrule
Transformer &          54.8\% &  58\%        & \textbf{61.1\%}      \\

\bottomrule
\end{tabular}
\end{table}

We also use log Mel-spectrogram for audio CNN, which obtains a little better performance, but the final results are very similar after intra- and cross-modal fusion. 
Besides, the concatenation of audio and visual vectors gets 58\% accuracy in AFEW validation set with transformer attention. This is 3\% lower than FBP which shows the effectiveness of FBP.

\subsection{Feature Enhancement}
In the Table~\ref{tab:feature_augmentation}, the \textbf{Basic Features} means that we only extract one feature vector for each frame. Besides, 
We apply 5 kinds of feature enhancement strategies as presented in Table~\ref{tab:feature_augmentation}. Specifically, for feature $F$-$Mean$, we first obtain 18 transformation frames by using three rotations, three scales, and flipping for a frame. After that, we compute the features of these 18 transformation frames and average these 18 features as the feature $F$-$Mean$. For the feature $F$-$MeanStd$, we compute the average feature and feature standard deviation of these 18 features. We then concatenate the average feature and the standard deviation as $F$-$MeanStd$. For the feature $F$-$normFFT$, we first compute the Fast Fourier transform(FFT) of the Basic Feature, and then normalize the feature and concatenate the real and imaginary parts as $F$-$normFFT$. 
For the feature  $F$-$AR$-$Mean$, $A$ means that the features are extracted by the models pre-trained on Affectnet, and $R$ by the models pre-trained on RAF-DB. we concatenate these two mean features of two different pretrained models as $F$-$AR$-$Mean$.

\begin{table}[htp]
\footnotesize
\centering
\newcommand{\tabincell}[2]{\begin{tabular}{@{}#1@{}}#2\end{tabular}}
\caption{Evaluation of five feature enhancement strategies. The default setting is Rotation $\in [-2\degree, 0\degree, 2\degree], \textbf{scale} \in [1, 1.03, 1.07]$ }
\label{tab:feature_augmentation}
\begin{tabular}{ccc}
\toprule
Visual Feature  & Augmentation details &  AFEW Val acc                \\ \midrule
\textbf{Basic Feature}    &  ----    &    \textbf{61.1\%  }        \\ 
\textbf{Basic Feature\_RAF-DB}    &  ----    &    58.5\%          \\ \midrule
F-Mean   & \emph{default setting} &   62.14\%         \\ 

F-MeanStd    & \emph{default setting} &   \textbf{63.7\%}         \\

F-MeanStd-2    & \tabincell{c}{ $Rotation \in [-15\degree,0\degree,15\degree]$\\ $scale \in [0.75, 1, 1.25] $  } &   62.4\%         \\

F-NormFFT    &  Normalized FFT  &   61.35\%         \\

F-AR-Mean    & \emph{default setting} &   62.92\%         \\

\midrule
FG-Net   & ----  & 59\%  \\
\bottomrule
\end{tabular}
\end{table}

Table ~\ref{tab:feature_augmentation} shows that the five feature enhancement methods further improve the performance of FBP where the feature F-MeanStd achieves the best result on the validation set. 

\begin{table}[htp]
\footnotesize
\newcommand{\tabincell}[2]{\begin{tabular}{@{}#1@{}}#2\end{tabular}}
\centering
\caption{Submission results of different model combinations.}
\label{tab:submit_run}
\begin{tabular}{cccccc}
\toprule
Sub & Val & Test   & Fusion detail    \\ \midrule
(1)    &    ----       &  \textbf{62.481\%}\   &4~FG-Net-1 \\
(2)    &    ----        & 59.112\%  &  2~F-MeabStd-2 + 2~F-AR-Mean \\
(3)    &    ----      &  54.518\%  & 4~FG-Net-2\\            
(4)    &     64.5\%      & {\bfseries 61.41\%}  & 4~F-MeanStd  \\
(5)    &     65.5\%      & {\bfseries 62.328\%}  & \tabincell{c}{F-Mean + F-MeanStd + F-NormFFT \\+ F-MeanStd-2 + F-AR-Meam}  \\  \bottomrule
\end{tabular}
\end{table}

\subsection{Results On EmotiW2019}
In the Table \ref{tab:submit_run}. The first three submitted models are trained on the training and validation set of AFEW, and the last two models are trained on the training set of AFEW. We find that it is difficult to choose models and fuse models if combining the validation set with the training set. We adopt class weight in all submissions, which means that we re-weight the predicted scores by the square root of the sample numbers([0.15, 0.097, 0.129, 0.185, 0.138, 0.082, 0.215]).

\section{Conclusions}
In this paper, we exploit three types of intra-modal fusion methods, namely self-attention, relation-attention, and transformer. They are mainly used to highlight important emotion feature. For the fusion of audio and visual information, we explore feature concatenation and factorized bilinear pooling (FBP). 
Besides, we evaluate different emotion features, including an audio feature with both speech-spectrogram and Log Mel-spectrogram and several facial features with different CNN models and different emotion pretrained strategies. With careful evaluation, we obtain 62.48\% and rank second in the EmotiW 2019 Challenge.

\section{Acknowledgments}
This work is partially supported by the National Natural Science Foundation of China (U1613211), Shenzhen Basic Research Program (JCYJ20170818164704758), the Joint Lab of CAS-HK.

\bibliographystyle{ACM-Reference-Format}
\bibliography{acmart}

\end{document}